%% file: main.tex
\documentclass[10pt,twocolumn,letterpaper]{article}

\usepackage{cvpr}              
\usepackage{marvosym}
\input{preamble}

\usepackage[accsupp]{axessibility} 

\definecolor{cvprblue}{rgb}{0.21,0.49,0.74}
\usepackage[pagebackref,breaklinks,colorlinks,citecolor=cvprblue]{hyperref}
\usepackage{multirow}
\usepackage{colortbl}

\def\paperID{3473} 
\def\confName{CVPR}
\def\confYear{2024}

\title{360DVD: Controllable Panorama Video Generation with \\ 360-Degree Video Diffusion Model}

\author{Qian Wang\textsuperscript{1,2}, Weiqi Li\textsuperscript{1}, Chong Mou\textsuperscript{1,2}, Xinhua Cheng$^{1,2}$, Jian Zhang\textsuperscript{1,2 \Letter}\\
\textsuperscript{1}School of Electronic and Computer Engineering, Peking University\\
\textsuperscript{2}Peking University Shenzhen Graduate School-Rabbitpre AIGC Joint Research Laboratory\\
{\tt\small \{qianwang,liweiqi,eechongm,chengxinhua\}@stu.pku.edu.cn, zhangjian.sz@pku.edu.cn}\\
}

\begin{document}


\maketitle

\let\thefootnote\relax\footnotetext{
This work was supported by National Natural Science Foundation of China under Grant 62372016. (\text{\Letter} Corresponding author: Jian Zhang)
}
\let\thefootnote\relax\footnotetext{
The authors would like to thank Miguel Lara (reachable at miguel@deforum.art) and Rahmel Jackson (reachable at rahmelsjackson@gmail.com), for their contributions to remarkable LatentLabs360.
}

\input{sec/0_abstract}    
\input{sec/1_intro}

\input{sec/2_relate}
\input{sec/3_method}
\input{sec/4_exp}
\input{sec/5_conclu}
{
    \small
    \bibliographystyle{ieeenat_fullname}
    \bibliography{main}
}


\end{document}

%% file: preamble.tex
\usepackage[dvipsnames]{xcolor}


%% file: sec/0_abstract.tex
\begin{abstract}
Panorama video recently attracts more interest in both study and application, courtesy of its immersive experience. 
Due to the expensive cost of capturing 360$^{\circ}$ panoramic videos, generating desirable panorama videos by prompts is urgently required. 
Lately, the emerging text-to-video (T2V) diffusion methods  
demonstrate notable effectiveness in standard video generation. 
However, due to the significant gap in content and motion patterns between panoramic and standard videos, these methods encounter challenges in yielding satisfactory 360$^{\circ}$ panoramic videos. 
In this paper, we propose a pipeline named \textbf{360}-\textbf{D}egree \textbf{V}ideo \textbf{D}iffusion model (\textbf{360DVD}) for generating 360$^{\circ}$ panoramic videos based on the given prompts and motion conditions.
Specifically, we introduce a lightweight 360-Adapter accompanied by 360 Enhancement Techniques to transform pre-trained T2V models for panorama video generation. 
We further propose a new panorama dataset named WEB360 consisting of panoramic video-text pairs for training 360DVD, addressing the absence of captioned panoramic video datasets. 
Extensive experiments demonstrate the superiority and effectiveness of 360DVD for panorama video generation.
Our project page is at \url{https://akaneqwq.github.io/360DVD/}.

\end{abstract}

\begin{figure*}[t]
    \centering
    \captionsetup{type=figure}
    \includegraphics[width=1.0\linewidth]{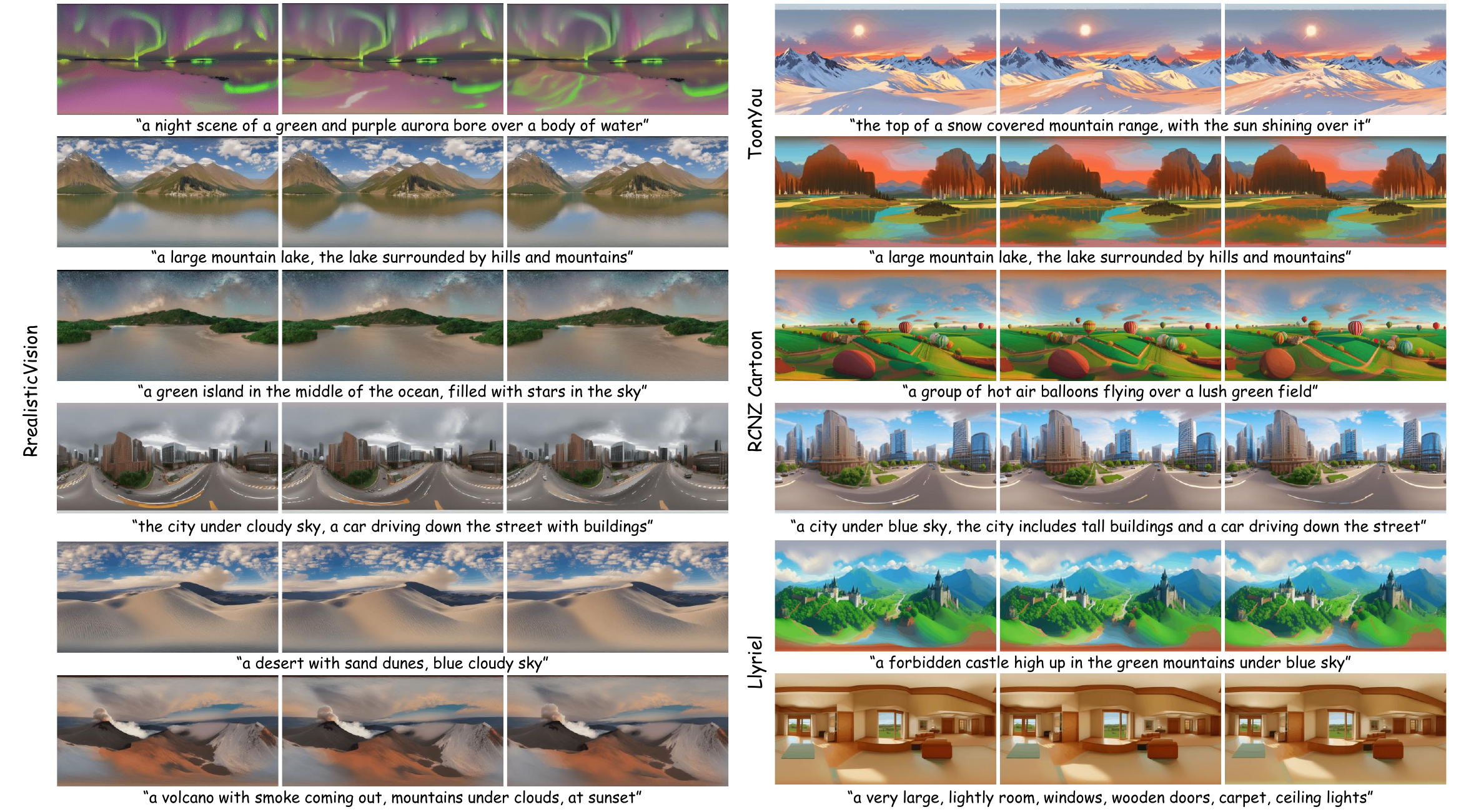}
    \captionof{figure}{\textbf{Main results.} Our 360DVD creates text-aligned, coherent, and high-quality $360^{\circ}$ panorama videos.
    Furthermore, 360DVD can cooperate with multiple personalized text-to-image models and consistently generate stylized panorama videos.}
    \label{fig:main}
\end{figure*}


%% file: sec/1_intro.tex
\section{Introduction}
\label{sec:intro}
With the recent advancements in VR technology, 360-degree panoramic videos have been gaining increasing popularity. This video format which offers audiences an immersive experience, is helpful for various applications, including entertainment, education, and communication. To capture details of the entire scene, $360^{\circ}$ videos are typically recorded using an array of high-resolution fisheye cameras that yields a $360^{\circ} \times 180^{\circ}$ field-of-view (FoV)~\cite{Omnidirectional_Vision_survey}, which is quite costly in both time and resources. Therefore, the generation of $360^{\circ}$ panoramic videos is urgently required for border applications, while panoramic video generation receives little attention in studies to date. 

Thanks to the emerging theory and training strategies,
text-to-image (T2I) diffusion models~\cite{Photorealistic, High-Resolution, Hierarchical, GLIDE, t2i-adapter} demonstrate remarkable image generation capacity from prompts given by users, and such impressive achievement in image generation is further extended to text-to-video (T2V) generation. 
Various T2V diffusion models~\cite{Make-A-Video, MagicVideo, Latent-Shift, videofactory, make-your-video, AnimateDiff} are recently proposed with adopting space-time separable architectures, wherein spatial operations are inherited from the pre-trained T2I models to reduce the complexity of constructing space-time models from scratch. 
Among these, AnimateDiff~\cite{AnimateDiff} enables the capability to generate animated images for various personalized T2I models, which alleviates the requirement for model-specific tuning and achieves compelling content consistency over time. 

Although T2V methods on standard videos are widely studied, there is no method proposed for panorama video generation.
One potential approach is to leverage existing powerful T2V models, \textit{e.g.}, AnimateDiff to directly generate the equirectangular projection (ERP) of panoramic videos. 
Since ERP is a commonly adopted format for storing and transmitting panoramic videos, each frame is treated by ERP as a rectangular image with an aspect ratio of 1:2, which aligns well with the output format of existing standard T2V models.
However, due to the significant differences between panoramic videos and standard videos, existing methods suffer challenges in directly producing satisfactory $360^\circ$ panoramic videos. Concretely, the main challenges include three aspects:
\textbf{(1)} The content distribution of ERPs differs from standard videos. ERPs require a wider FoV, reaching $360^{\circ} \times 180^{\circ}$. \textbf{(2)} The motion patterns of ERPs are different from standard videos, with movements often following curves rather than straight lines. \textbf{(3)} The left and right ends of ERPs should exhibit continuity since they correspond to the same meridian on the Earth.

Therefore, we propose a specifically designed method named \textbf{360}-\textbf{D}egree \textbf{V}ideo \textbf{D}iffusion (360DVD) for generating panorama videos.
We first introduce a plug-and-play module named 360-Adapter to address challenge mentioned above. 
Our 360-Adapter receives zero values or motion conditions (\textit{e.g.,} optical flow) as input and outputs motion features, which are fed into the frozen denoising U-Net at different levels of the encoder. 
This transformation is aimed at converting the T2V model into a panoramic video generation without altering the foundational generative capabilities. 
In addition, we introduce 360 Enhancement Techniques including two mechanisms to enhance continuity at both ends of ERPs from both macro and micro perspectives, and a latitude-aware loss function for encouraging the model to focus more on low-latitude regions. 
Cooperated with carefully designed techniques, our 360DVD generates text-aligned, coherent, high-quality, 360$^\circ$ panorama videos with various styles, as shown in Fig.~\ref{fig:main}.

Furthermore, 
we collect a panorama dataset named WEB360 including ERP-formatted videos from the internet and games for training our method.
WEB360 involves approximately 2,000 video clips with each clip consisting of 100 frames. 
Considering the domain gap between panoramic and standard images, to enhance the accuracy and granularity of captions, we introduce a GPT-based 360 Text Fusion module for obtaining detailed captions. 
Our contributions can be summarized as follows:
\begin{itemize}
    \item We introduce a controllable $360^{\circ}$ panorama video generation diffusion model named 360DVD, achieved by adopting a controllable standard T2V model with a trainable lightweight 360-Adapter. 
    Our model can generate text-guided panorama videos conditioned on desired motions.
    \item We design 360 Enhancement Techniques including a latitude-aware loss and two mechanisms to enhance the content and motion quality of generated panorama videos.
    \item We propose a new high-quality dataset named WEB360 comprising approximately 2,000 panoramic videos, with each video accompanied by a detailed caption enhanced through 360 Text Fusion.
    \item Experiments demonstrate that our 360DVD is capable of generating high-quality, high-diversity, and more consistent $360^{\circ}$ panorama videos.
\end{itemize}

%% file: sec/2_relate.tex
\section{Related Works}
\label{sec:relate}
\subsection{Text-to-Image Diffusion Model}
The Denoising Diffusion Probabilistic Model~\cite{ddpm, ddim, cheng2023null} has proven to be highly successful in generating high-quality images, outperforming previous approaches such as generative adversarial networks (GANs)\cite{gan, StackGAN}, variational autoencoders (VAEs)\cite{vae, vae2}, and flow-based methods~\cite{flow}. With text guidance during training, users can generate images based on textual input. Noteworthy examples include GLIDE~\cite{GLIDE}, DALLE-2~\cite{Hierarchical}, Imagen~\cite{Photorealistic}. To address the computational burden of the iterative denoising process, LDM~\cite{High-Resolution} conducts the diffusion process on a compressed latent space rather than the original pixel space. This accomplishment has prompted further exploration in extending customization~\cite{dreambooth, inversion}, image guidance~\cite{paint,yu2023freedom}, precise control~\cite{control-net, t2i-adapter, mou2023dragondiffusion} and protection~\cite{yu2023cross}.

\subsection{Text-to-Video Diffusion Model}
Despite significant advancements in Text-to-Image (T2I) generation, Text-to-Video (T2V) generation faces challenges, including the absence of large-scale, high-quality paired text-video datasets, the inherent complexity in modeling temporal consistency, and the resource-intensive nature of training. To address these challenges, many works leverage the knowledge from pre-trained T2I models, and they manage training costs by executing the diffusion process in the latent space. Some methods~\cite{tune-a-video, fatezero, Rerender-A-Video, TokenFlow, LAMP} utilize T2I models in zero-shot or few-shot ways. However, these methods often suffer from suboptimal frame consistency due to insufficient training. To address this limitation, another category of T2V diffusion models typically adopts space-time separable architectures. These models~\cite{Make-A-Video, MagicVideo, Latent-Shift, videofactory} inherit spatial operations from pre-trained T2I models, reducing the complexity of constructing space-time models from scratch. Given that most personalized T2I models are derived from the same base one (e.g. Stable Diffusion~\cite{High-Resolution}), AnimateDiff~\cite{AnimateDiff} designs a motion modeling module that trained with a base T2I model and could animate most of derived personalized T2I models once for all. There are also efforts focused on enhancing control in T2V models. Gen-1~\cite{gen1}, MCDiff~\cite{mcDiff}, LaMD~\cite{LaMD} and VideoComposer~\cite{VideoComposer} introduce diverse conditions to T2V models. Despite these advancements, the aforementioned methods demand extensive training and lack a plug-and-play nature, making it challenging to apply them to a diverse range of personalized T2I models.

\subsection{Panorama Generation}
GAN-based methods for generating panoramic images have been widely studied~\cite{oh2022bips, teterwak2019boundless, wu2022cross, lin2019coco, lin2021infinitygan, cheng2022inout, wang2022stylelight, chen2022text2light, dastjerdi2022guided, akimoto2022diverse, Cao_2023_CVPR, sun2023opdn}. For instance, OmniDreamer~\cite{akimoto2022diverse} accepts a single NFoV image as an input condition and introduces a cyclic inference scheme to meet the inherent horizontal cyclicity of 360-degree images. ImmenseGAN~\cite{dastjerdi2022guided} fine-tunes the generative model using a large-scale private text-image pair dataset, making the generation more controllable. Text2Light~\cite{chen2022text2light} introduces a zero-shot text-guided 360-image synthesis pipeline by utilizing the CLIP model. Very recently, diffusion models have achieved promising results in panoramic image generation. DiffCollage~\cite{zhang2023diffcollage} uses semantic maps as conditions and generates images based on complex factor graphs using retrained diffusion models. PanoGen~\cite{li2023panogen} employs a latent diffusion model and synthesizes new indoor panoramic images through recursive image drawing techniques based on multiple text descriptions. PanoDiff~\cite{panodiff} achieves a multi-NFoV synthesis of panoramic images through a two-stage pose estimation module. IPO-LDM~\cite{wu2023ipo} uses a dual-modal diffusion structure of RGB-D to better learn the spatial distribution and patterns of panoramic images. StitchDiffusion~\cite{wang2023customizing} employs a T2I diffusion model, ensuring continuity at both ends through stitching. However, to date, panoramic video generation has received limited attention. To the best of our knowledge, we are the first to leverage diffusion models for panoramic video generation.

%% file: sec/3_method.tex
\section{Method}
\label{method}
In this section, we begin with a concise review of the latent diffusion fusion model and AnimateDiff~\cite{AnimateDiff}. Following that, we introduce the construction method of the WEB360 dataset. We then provide an overview of 360DVD and elaborate on the implementation details of 360-Adapter. Finally, we describe the 360 enhancement techniques aimed at enriching the panoramic nature of the video.

\subsection{Preliminaries}
\noindent\textbf{Latent Diffusion Model.}
Given an input signal $\mathbf{x}_{0}$, a diffusion forward process in DDPM~\cite{ddpm} is defined as:
\begin{equation}
    p_\theta(\mathbf{x}_{t} | \mathbf{x}_{t-1}) = \mathcal{N}(\mathbf{x}_t; \sqrt{1 - \beta_{t}}\mathbf{x}_{t-1}, \beta_{t}\mathbf{I}),
\end{equation}
for $t = 1, \dotsc, T$, where $T$ is the total timestep of the diffusion process. A noise depending on the variance $\beta_{t}$ is gradually added to $\mathbf{x}_{t-1}$ to obtain $\mathbf{x}_t$ at the next timestep and finally reach $\mathbf{x}_T \in \mathcal{N}(0, \mathbf{I})$. The goal of the diffusion model is to learn to reverse the diffusion process (denoising). Given a random noise $\mathbf{x}_t$, the model predicts the added noise at the next timestep $\mathbf{x}_{t-1}$ until the origin signal $\mathbf{x}_0$:
\begin{equation}
    p_\theta(\mathbf{x}_{t-1} | \mathbf{x}_t) = \mathcal{N}(\mathbf{x}_{t-1}; {\boldsymbol{\mu}}_\theta(\mathbf{x}_t, t), {\boldsymbol{\Sigma}}_\theta(\mathbf{x}_t, t)),
\end{equation}
for $t = T, \dotsc, 1$. We fix the variance $\boldsymbol{\Sigma}_\theta(\mathbf{x}_t, t)$ and utilize the diffusion model with parameter $\theta$ to predict the mean of the inverse process ${\boldsymbol{\mu}}_\theta(\mathbf{x}_t, t)$. The model can be simplified as denoising models ${\boldsymbol{\epsilon}}_\theta(\mathbf{x}_t, t)$, which are trained to predict the noise of $\mathbf{x}_t$ with a noise prediction loss:
\begin{equation}
    \mathcal{L} = \mathbb{E}_{\mathbf{x}_0, \mathbf{y}, {\boldsymbol{\epsilon}} \sim \mathcal{N}(0, \mathbf{I}), t}[\left\|{\boldsymbol{\epsilon}} - {\boldsymbol{\epsilon}}_\theta(\mathbf{x}_t, t, \boldsymbol{\tau}_\theta(\mathbf{y}))\right\|^2_2],
\end{equation}
where ${\boldsymbol{\epsilon}}$ is the added noise to the input image $\mathbf{x}_0$, $\mathbf{y}$ is the corresponding textual description, $\boldsymbol{\tau}_\theta(\cdot)$ is a text encoder mapping the string to a sequence of vectors.

Latent Diffusion Model (LDM)~\cite{High-Resolution} executes the denoising process in the latent space of an autoencoder, namely $\mathcal{E}(\cdot)$ and $\mathcal{D}(\cdot)$, implemented as VQ-GAN~\cite{VQ-GAN} or VQ-VAE~\cite{VQ-VAE} pre-trained on large image datasets. During the training of the latent diffusion networks, an input image $\mathbf{x}_0$ is initially mapped to the latent space by the frozen encoder, yielding $\mathbf{z}_0 = \mathcal{E}(\mathbf{x}_0)$. Thus, the training objective can be formulated as follows:
\begin{equation}
    \mathcal{L} = \mathbb{E}_{\mathcal{E}(\mathbf{x}_0), \mathbf{y}, {\boldsymbol{\epsilon}} \sim \mathcal{N}(0, \mathbf{I}), t}[\left\|{\boldsymbol{\epsilon}} - {\boldsymbol{\epsilon}}_\theta(\mathbf{z}_t, t, \boldsymbol{\tau}_\theta(\mathbf{y}))\right\|^2_2].
\end{equation}

In widely-used LDM Stable Diffusion (SD), which our method is based on, ${\boldsymbol{\epsilon}}_\theta(\cdot)$ is implemented with a modified UNet~\cite{unet} that incorporates four downsample/upsample blocks and one middle block, resulting in four resolution levels within the networks' latent space. Each resolution level integrates 2D convolution layers as well as self- and cross-attention mechanisms. Text model $\boldsymbol{\tau}_\theta(\cdot)$ is implemented using the CLIP~\cite{clip} ViT-L/14 text encoder.

\noindent\textbf{AnimateDiff.} AnimateDiff inflates base SD by adding temporal-aware structures and learning reasonable motion priors from large-scale video datasets. 
Since the original SD can only process 4D image data batches, while T2V task takes a 5D video tensor as input. It transforms each 2D convolution and attention layer in the original image model into spatial-only pseudo-3D layers. 
The motion module is inserted at every resolution level of the U-shaped diffusion network, using vanilla temporal transformers consisting of several self-attention blocks operating along the temporal axis. The training objective of AnimateDiff can be written as:
\begin{equation}
    \mathcal{L} = \mathbb{E}_{\mathcal{E}(\mathbf{x}_0^{1:N}), \mathbf{y}, {\boldsymbol{\epsilon}} \sim \mathcal{N}(0, \mathbf{I}), t}[||{\boldsymbol{\epsilon}} - {\boldsymbol{\epsilon}}_\theta(\mathbf{z}_t^{1:N}, t, \boldsymbol{\tau}_\theta(\mathbf{y}))||^2_2],
\end{equation}
where $\mathbf{x}_0^{1:N}$ is the sampled video data, $\mathbf{z}_0^{1:N}$ is the latent code which $\mathbf{x}_0^{1:N}$ are encoded into via the pre-trained autoencoder, $\mathbf{z}_t^{1:N}$ is the latent code obtained by perturbing the initial latent code $\mathbf{z}_0^{1:N}$ with noise at timestep $t$. During training, the pre-trained weights of the base T2I model are frozen to keep its feature space unchanged.


\subsection{WEB360 Dataset}
Diverse text-video pairs datasets are essential for training open-domain text-to-video generation models. However, existing $360^{\circ}$ panorama video datasets lack corresponding textual annotations. Moreover, these datasets are often constrained either in scale or quality, thereby impeding the upper limit of high-quality video generation. 

\begin{figure}[t]
    \centering
    \captionsetup{type=figure}
    \includegraphics[width=1.0\linewidth]{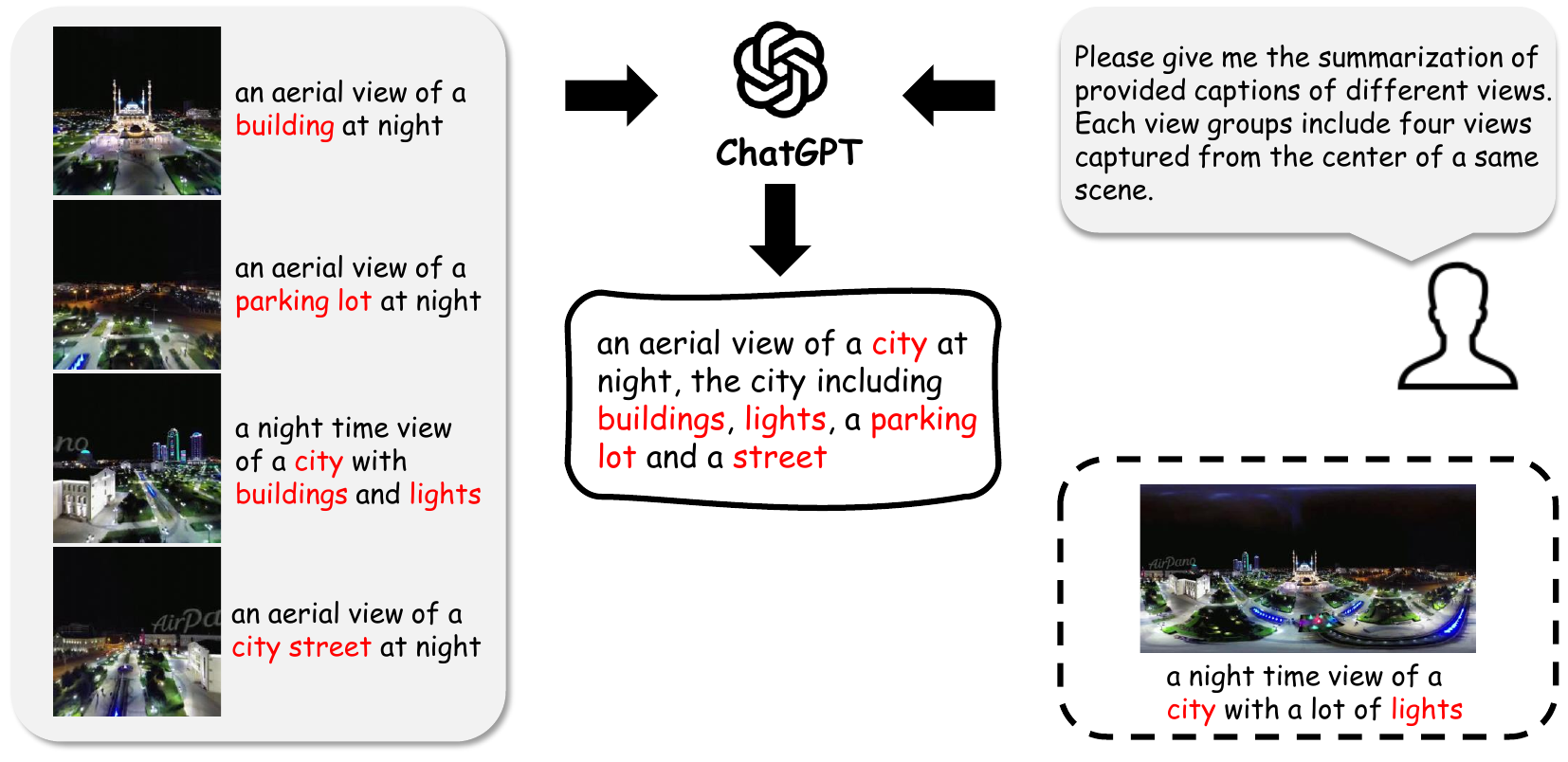}
    \captionof{figure}{\textbf{360 Text Fusion.} The captions of four images with a FoV of 90 are fed into ChatGPT to generate a new $360^{\circ}$ summarization. Compared to the caption of ERP at the bottom right, 360 Text Fusion allows for more fine-grained captions.}
    \label{fig:360tf}
\end{figure}

To address the aforementioned challenges and achieve high-quality 360 panorama video generation, we introduce a novel text-video dataset named WEB360. This dataset comprises 2114 text-video pairs sourced from open-domain content, presented in high-definition (720p) ERP format. Our dataset creation process involved extracting 210 high-resolution panoramic video clips from the ODV360~\cite{Cao_2023_CVPR} training set. Additionally, we collected over 400 original videos from YouTube. Due to the complex scene transitions present in the original videos, which pose challenges for models in learning temporal correlations, we perform a manual screening process to split the original videos into 1904 single-scene video clips. We employ BLIP~\cite{blip} to annotate the first frame of the 2104 video clips. However, we observed that direct application of BLIP to ERP images often resulted in bad captions. Therefore, we propose a panoramic image caption method named 360 Text Fusion, based on ChatGPT. 

\begin{figure*}[t]
    \centering
    \captionsetup{type=figure}
    \includegraphics[width=1.0\linewidth]{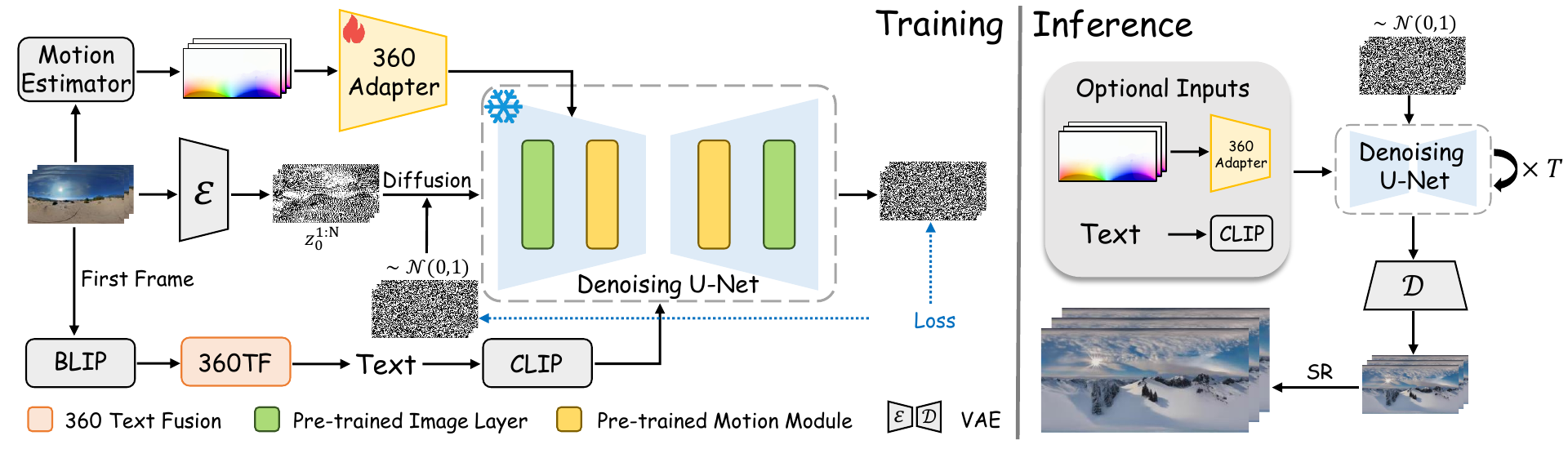}
    \captionof{figure}{\textbf{Overview of 360DVD.}
    360DVD leverages a trainable 360-Adapter to extend standard T2V models to the panorama domain and is able to generate high-quality panorama videos with given prompts and optional motion conditions.
    In addition, 360 Enhancement Techniques are proposed for quality improvement in the panorama perspective. 
    }
    \label{fig:pipeline}
\end{figure*}

\noindent\textbf{360 Text Fusion.} We find that directly using BLIP~\cite{blip} to label ERP has drawbacks. On one hand, errors may arise due to the distortion caused by the polarities, leading to misidentifications such as labeling ``person" as ``dog". On the other hand, the captions generated by BLIP lack granularity, making them insufficient for providing a detailed description of the current scene. Thus, we propose 360 Text Fusion (360TF) method, as shown in Fig.~\ref{fig:360tf}. To deal with the irregular distortion of ERP, we turn to less-distorted perspective images. We first project the original ERP image to four non-overlapping perspective images at 0 degrees longitude, with a FoV of 90. The four images are then fed into BLIP to be captioned. By pre-informing ChatGPT about the task and providing examples, these four captions are collectively input to ChatGPT, which then generates a summary of the scene as our final caption. In comparison to directly using BLIP to label the entire image, our 360TF demonstrates a significant advantage in granularity.

\subsection{360-degree Video Diffusion Model}
An overview of the 360-degree Video Diffusion Model (360 DVD) is presented in Fig.~\ref{fig:pipeline}, which is composed of a pre-trained denoising U-Net and 360-Adapter. The pre-trained denoising U-Net adopts a structure identical to that of AnimateDiff. In every resolution level of the U-Net, the spatial layer unfolds pre-trained weights from SD, while the temporal layer incorporates the motion module of AnimateDiff trained on a large-scale text-video dataset. 

During the training process, we first sample a video $\mathbf{x}_0^{1:N}$ from the dataset. The video is encoded into latent code $\mathbf{z}_0^{1:N}$ through pre-trained VAE encoder $\mathcal{E}(\cdot)$ and noised to $\mathbf{z}_t^{1:N}$. Simultaneously, the corresponding text $\mathbf{y}$ for the video is encoded using the text encoder $\boldsymbol{\tau}_\theta(\cdot)$ of the CLIP. The video is also input into a motion estimation network to generate corresponding motion conditions $\mathbf{c}$, which are then fed into the 360-Adapter $\mathcal{F}_{360}(\cdot)$. Finally, noised latent code $\mathbf{z}_t^{1:N}$, timestep $t$, text embedding $\boldsymbol{\tau}_{\theta}(\mathbf{y})$, and the feature maps $\mathbf{f}_{360}$ generated by 360-Adapter are collectively input into the U-Net ${\boldsymbol{\epsilon}}(\cdot)$ to predict the noise strength added to the latent code. As we aim to preserve the priors learned by SD and AnimateDiff on large datasets, we freeze their weights during the training process. If we use a simple L2 loss term, the training objective is given as follows:
\begin{equation}
    \small
    \mathcal{L} = \mathbb{E}_{\mathcal{E}(\mathbf{x}_0^{1:N}), \mathbf{y}, {\boldsymbol{\epsilon}} \sim \mathcal{N}(0, \mathbf{I}), t}[||{\boldsymbol{\epsilon}} - {\boldsymbol{\epsilon}}_\theta(\mathbf{z}_t^{1:N}, t, \boldsymbol{\tau}_\theta(\mathbf{y}), \mathbf{f}_{360})||^2_2].
\end{equation}

To ensure satisfactory generation of $360^{\circ}$ panoramic videos without motion control input, we set the input of the 360-Adapter to zero with a probability $P$ during training. This strategy aims to encourage the model to learn representations that are not solely reliant on motion conditions, enhancing its ability to generate compelling panoramic videos without explicit motion guidance.

\begin{figure}[t]
    \centering
    \includegraphics[width=0.85\linewidth]{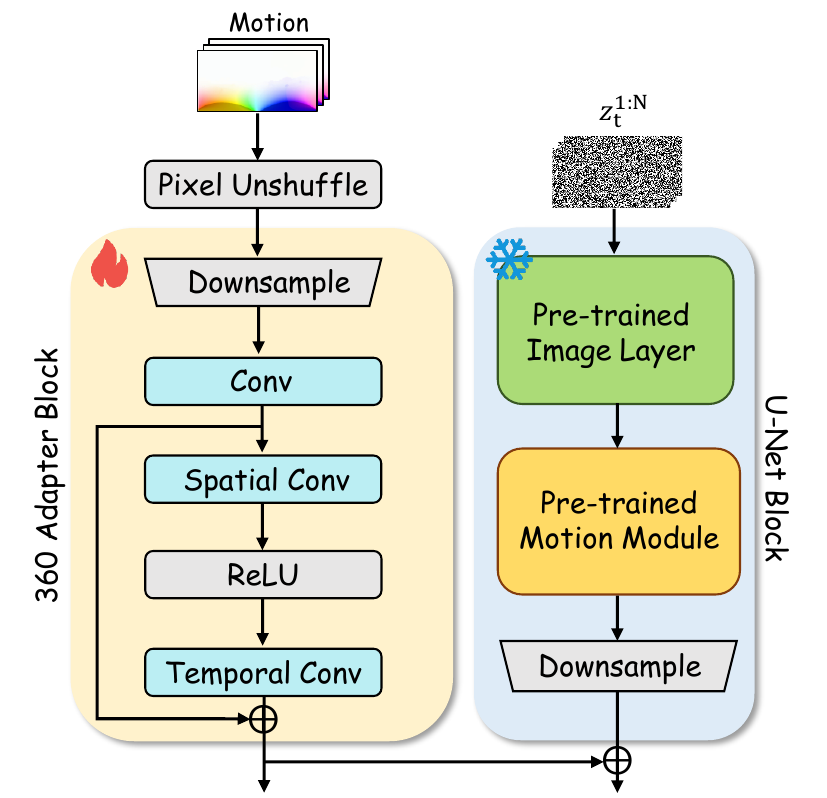}
    \caption{\textbf{Overview of 360-Adapter.} 360-Adapter is a simple but effective module in which intermediate features are fed into the U-Net encoder blocks for modulation.}
    \label{fig:360Adapter}
\end{figure}

In inference, users have the option to selectively provide text prompts and motion guidance to carry out denoising over a total of $T$ steps. Here, we employ DDIM~\cite{ddim} to accelerate the sampling process. The estimated latent code $\hat{\mathbf{z}}_0^{1:N}$ is then input into a pre-trained VAE decoder to decode the desired $360^{\circ}$ panoramic videos $\hat{\mathbf{x}}_0^{1:N}$. Due to constraints such as resolution limitations imposed by existing SD and considerations regarding GPU memory usage, the experimental results presented in this paper showcase a resolution of $512\times1024$. In practical applications, super-resolution methods~\cite{sun2023opdn, cheng2023hybrid} can be employed to upscale the generated results to the desired size.

\noindent\textbf{360-Adapter.}
Our proposed 360-Adapter is simple and lightweight, as shown in Fig.~\ref{fig:360Adapter}. The original condition input has the the same resolution as the video of $H \times W$. Here, we utilize the pixel unshuffle~\cite{pixelunshuffle} operation to downsample it to $H/8 \times W/8$. Following that are four 360-Adapter blocks, we depict only one for simplification in Fig.~\ref{fig:360Adapter}. To maintain consistency with the U-Net architecture, the first three 360-Adapter blocks each include a downsampling block. In each 360-Adapter block, one 2D convolution layer and a residual block (RB) with pseudo-3D convolution layers are utilized to extract the condition feature $\mathbf{f}^k_{360}$. Finally, multi-scale condition features $\mathbf{f}_{360} = \{\mathbf{f}^1_{360} , \mathbf{f}^2_{360}, \mathbf{f}^3_{360}, \mathbf{f}^4_{360} \}$ are formed. Suppose the intermediate features in the U-Net encoder block is $\mathbf{f}_{enc} = \{\mathbf{f}^1_{enc}, \mathbf{f}^2_{enc}, \mathbf{f}^3_{enc}, \mathbf{f}^4_{enc} \}$. $\mathbf{f}_{360}$ is then added with $\mathbf{f}_{enc}$ at each scale. In summary, the condition feature extraction and conditioning operation of the 360-Adapter can be defined as the following formulation:
\begin{equation}
    \mathbf{f}_{360} = \mathcal{F}_{360}(\mathbf{c}),
\end{equation}
\begin{equation}
    \hat{\mathbf{f}}^i_{enc} = \mathbf{f}^i_{enc} + \mathbf{f}^i_{360}, i \in \{1, 2, 3, 4\}.
\end{equation}

In the previous description, we omit some details. Our motion condition $\mathbf{c}$ is a 5D tensor, assuming its size is $batch \times channels \times frames \times height \times width$. We first reshape it into a 4D tensor of size $(batch \times frames) \times channels \times height \times width$ to allow it to be fed into the 2D convolution layer and restore it to 5D to go through the RB with pseudo-3D convolution layers. Subsequently, in the RB, we employ a $1 \times 3 \times 3$ pseudo-3D convolution to extract features in the spatial dimension, followed by a $3 \times 1 \times 1$ pseudo-3D convolution to model information along the temporal dimension. The resulting features are reshaped back to $(batch \times frames) \times channels \times height \times width$ to add the output of the skip connection. Finally, condition features are reshaped back into a 5D vector of size $batch \times channels \times frames \times height \times width$ to align with the U-Net encoder intermediate features.

\subsection{360 Enhancement Techniques}
\begin{figure}[t]
    \centering
    \includegraphics[width=1.0\linewidth]{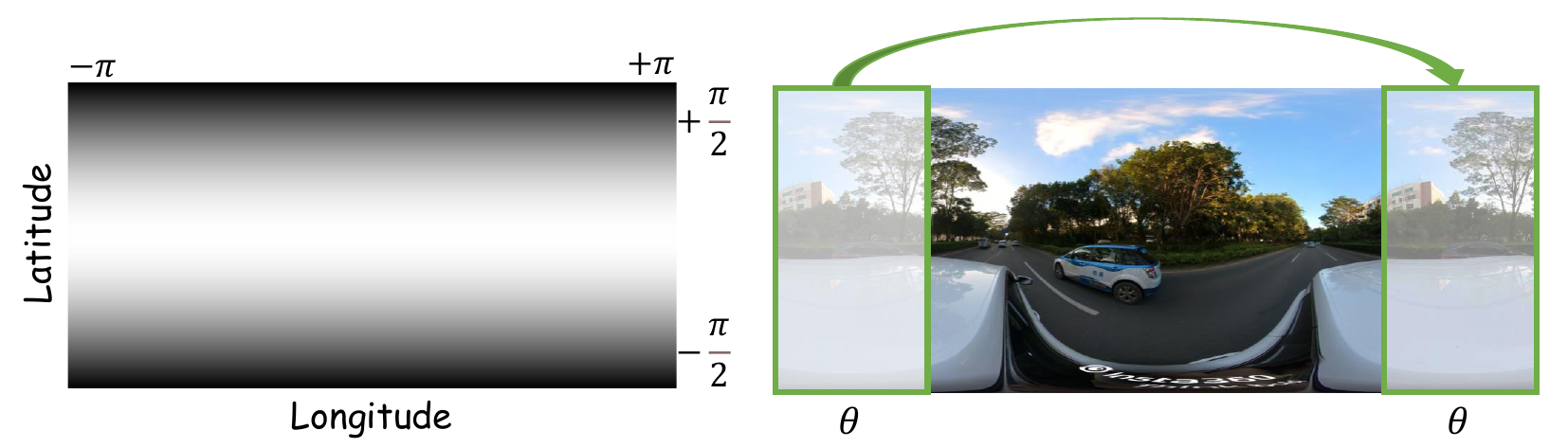}
    \caption{\textbf{Left:} the visualization of weight matrix $\mathbf{W}$, brighter colors indicate values closer to 1, while darker colors suggest values closer to 0. \textbf{Right:} a schematic diagram of the latent rotation mechanism. In each iteration, the far left portion of angle $\theta$ is shifted to the far right.}
    \label{fig:latitude}
\end{figure}

\noindent\textbf{Latitude-aware Loss.}
When projecting panoramic videos into ERPs, meridians are mapped as vertically spaced lines with a constant interval, while parallels are mapped as horizontally spaced lines with a constant interval. This projection method establishes a straightforward mapping relationship, but it is neither equal-area nor conformal, introducing significant distortion, particularly in the polar regions. To make the denoiser pay more attention to low-latitude regions with less distortion, which is more crucial for human visual perception, we introduce a latitude-aware loss:
\begin{equation}
\begin{aligned}
    \mathcal{L} = \mathbb{E}_{\mathcal{E}(\mathbf{x}_0^{1:N}), \mathbf{y}, {\boldsymbol{\epsilon}} \sim \mathcal{N}(0, \mathbf{I}), t}[||\mathbf{W} \odot({\boldsymbol{\epsilon}} - \hat{{\boldsymbol{\epsilon}}}_\theta)||^2_2],
\end{aligned}
\end{equation}
where $\hat{{\boldsymbol{\epsilon}}}_\theta={\boldsymbol{\epsilon}}_\theta(\mathbf{z}_t^{1:N}, t, \boldsymbol{\tau}_\theta(\mathbf{y}), \mathbf{f}_{360})$, and $\mathbf{W}$ is a weight matrix used to perform element-wise product, defined as:
\begin{equation}
    \mathbf{W}_{i,j} = \cos{(\frac{2i-H/8+1}{H/4}\pi)},
\end{equation}
where $i \in \left [0, H/8 \right )$, $j \in \left [0, W/8 \right)$, $H/8$ and $W/8$ is the height and width of latent code $\mathbf{z}_t^{1:N}$. The visualized result of $\mathbf{W}$ is shown in Fig.~\ref{fig:latitude}, where pixels in low and middle latitudes are given more weight during training.

\begin{figure*}[htbp]
    \centering
    \includegraphics[width=1.0\linewidth]{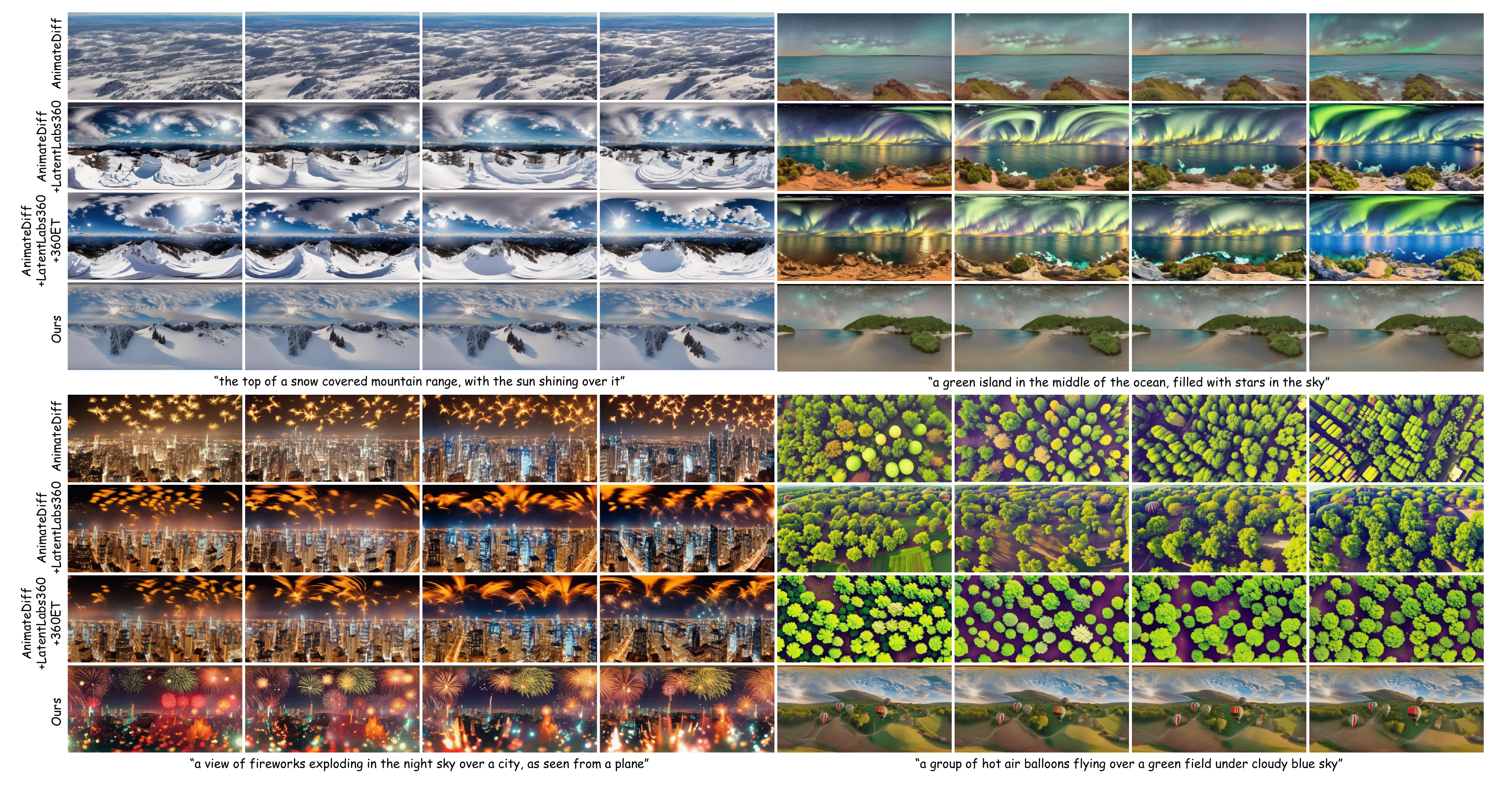}
    \caption{\textbf{Qualitative comparisons with baseline methods.} 360DVD successfully produces stable and high-quality panorama video over various prompts while other methods are failed.}
    \label{fig:comparison}
\end{figure*}

\begin{figure*}[htbp]
    \centering
    \includegraphics[width=1.0\linewidth]{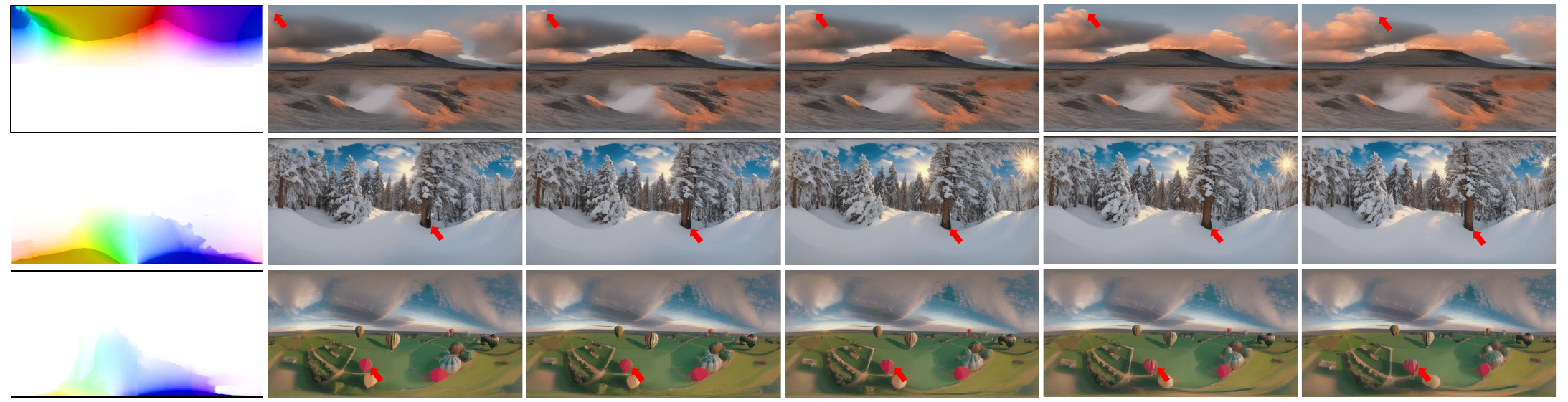}
    \caption{\textbf{Qualitative comparisons of optical flow.} 360DVD generates panorama videos with reasonable motion patterns consistent with the conditioned optical flow.}
    \label{fig:flow}
\end{figure*}

\noindent\textbf{Latent Rotation Mechanism.}
Because ERPs can be considered as the unfolding of a spherical surface along a meridian, they are meant to be wraparound consistent, implying that their left and right sides are continuous. However, during the process of video generation, the left and right sides are physically separated. Inspired by PanoDiff~\cite{panodiff}, we employ a latent rotation mechanism to enhance the macroscopic coherence between the left and right ends of the video. During the inference process, we perform a horizontal rotation at an angle of $\theta$ on $\mathbf{z}_t^{1:N}$ and motion condition $\mathbf{c}$, at each denoising step. As illustrated in Fig.~\ref{fig:latitude}, the content on the far left is shifted to the far right, where we use $\mathbf{x}_0^{1}$ to replace $\mathbf{z}_t^{1:N}$ for a better visual effect of its continuity. During the training process, we also randomly rotate the training videos along with the motion condition by a random angle as a data augmentation strategy.

\noindent\textbf{Circular Padding Mechanism.}
Although the previous latent rotation mechanism achieves semantic continuity at a macroscopic level, achieving pixel-level continuity is challenging. Therefore, in the inference process, we adopt a mechanism of circular padding by modifying the padding method of the convolution layers. We observe that the early stages of $360^{\circ}$ video generation often involve layout modeling, while the later stages focus on detail completion. To maintain the stable video generation quality of 360DVD, we only implement the circular padding mechanism in the late $\lfloor \frac{T}{2} \rfloor$ steps of a total of $T$ denoising steps.

%% file: sec/4_exp.tex
\section{Experiment}
\label{exp}

\begin{table*}
    \centering
    \setlength{\tabcolsep}{5pt}
    \begin{tabular}{ccccccc}
        \toprule
         & & \multicolumn{2}{c}{Video Criteria} & \multicolumn{3}{c}{Panorama Criteria} \\
         \cmidrule(r){3-4}\cmidrule(r){5-7}
        Index & Methods & Graphics Quality & Frame Consistency & End Continuity & Content Distribution & Motion Pattern\\
        \midrule
        A & AnimateDiff & 11.3\% & 15.3\% & 5.3\% & 4.8\% & 4.4\% \\
        B & A+LoRA & 14.1\% & 10.5\% & 6.0\% & 12.1\% & 6.5\% \\
        C &  B+360ET & 23.0\% & 9.7\% & 16.9\% & 16.1\% & 14.5\% \\
        \rowcolor{gray!20}
        D & Ours & 51.6\% & 64.5\% & 71.8\% & 67.0\% & 74.6\% \\
        \bottomrule
    \end{tabular}
    \caption{\textbf{User preference studies.} More raters prefer videos generated by our 360DVD, especially over panorama criteria including if generated videos have left-to-right continuity, the panorama content distribution, and the panorama motion pattern.}
    \label{tab:web360}

\end{table*}

\subsection{Implementation Details}
\noindent\textbf{Training Settings.}
We choose Stable Diffusion v1.5 and Motion Module v14 as our base model. We utilize the panoramic optical flow estimator PanoFlow~\cite{panodiff} to generate motion conditions. We train the 360-Adapter using the proposed WEB360 dataset. The resolution is set to $512 \times 1024$, the length of frames to $16$, the batch size to $1$, the learning rate to $1\times 10^{-5}$, and the total number of training steps to $100k$, probability $P = 0.2$. We use a linear beta schedule as AnimateDiff, where $\beta_{start} = 0.00085$ and $\beta_{end} = 0.012$. 

\noindent\textbf{Inference Settings.}
We use DDIM with 25 sampling steps, and the scale for text guidance is 7.5,  the angle $\theta = \pi/2 $. We collect several personalized Stable Diffusion models from CivitAI to verify the effectiveness and generalizability of our method, including Realistic Vision, Lyriel, ToonYou, and RCNZ Cartoon.

\subsection{Qualitative Results}
Due to space limitations, we only display several frames of each video. We strongly recommend readers refer to our project page for more results and better visual quality. 

\noindent\textbf{Prompt-guided Panorama Video Generation.}
We present several prompt-guided $360^{\circ}$ panorama video generation results across different personalized models in Fig.~\ref{fig:main}. The figure shows that our method successfully turns personalized T2I models into panorama video generators. Our method can produce impressive generation results ranging from real to cartoon styles, from natural landscapes to cultural scenery. This success is attributed to the fact that our method preserves the image generation priors and temporal modeling priors learned by SD and AnimateDiff on large-scale datasets.

\noindent\textbf{Motion-guided Panorama Video Generation.}
We showcase panoramic video generation results guided by three typical optical flow maps, as shown in Fig.~\ref{fig:flow}. The optical flow maps in the first row indicate the primary motion areas in the Arctic, where we can observe significant movement of clouds in the sky. The optical flow maps in the second row and third row indicate motion areas primarily in the Antarctic, where we can see the movement of trees and hot air balloons near the Antarctic.

\subsection{Comparison}
We compare our results with native AnimateDiff, AnimateDiff with a LoRA for panorama image generation from CivitAI named LatentLabs360, AnimateDiff with panoramic LoRA, and our proposed 360 Enhancement Techniques (loss excepted). We can observe that the results generated by the native AnimateDiff have a very narrow field of view, which does not align with the content distribution of panoramic videos. When AnimateDiff is augmented with panoramic LoRA, it produces videos with a broader field of view; however, the two ends of videos lack continuity, and object movements are highly random. Our proposed 360ET method significantly enhances the continuity between two ends of the videos but fails to address issues such as non-compliance with panoramic motion patterns and poor cross-frame consistency. Notably, our 360DVD can generate videos that best adhere to the content distribution and motion patterns of panoramic videos. We are pleased to discover that, thanks to the high-quality training data provided by WEB360, the videos generated by 360DVD exhibit more realistic colors and nuanced lighting, providing an immersive experience.

\subsection{Ablation Study}
\begin{figure}[t!]
    \centering
    \includegraphics[width=1.0\linewidth]{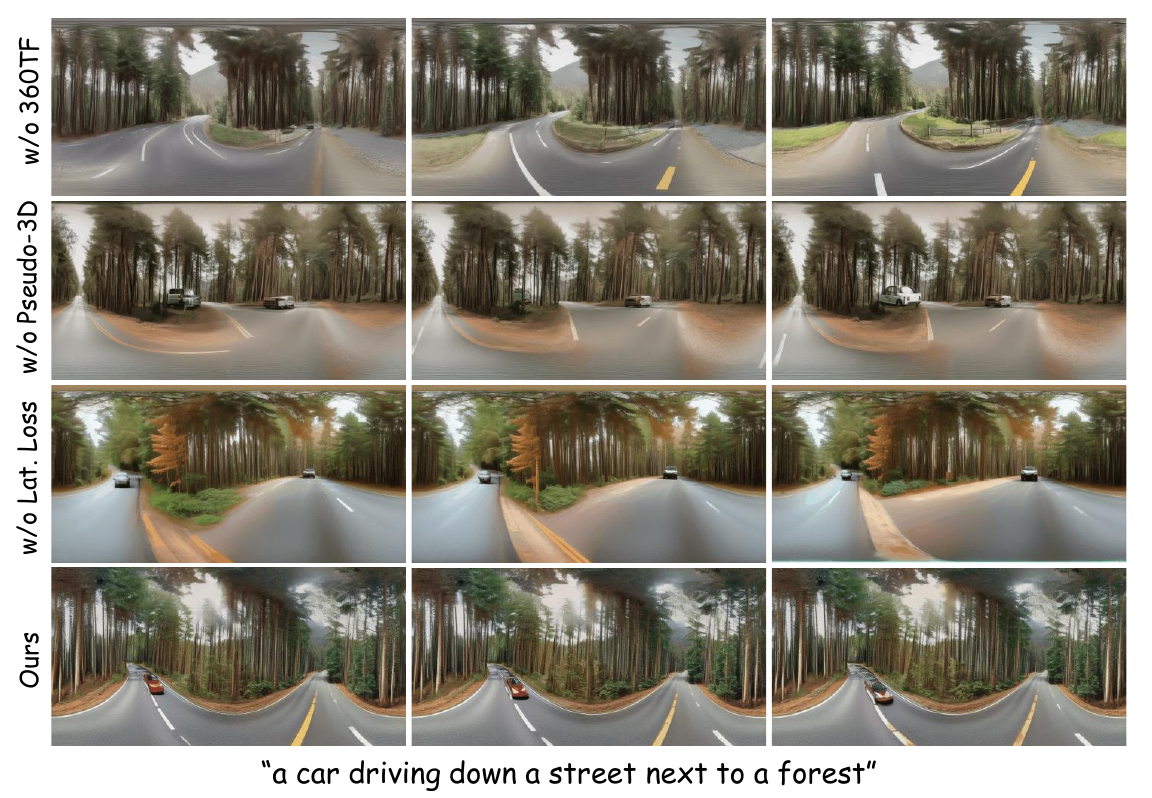}
    \caption{\textbf{Ablation studies} on 360 Text Fusion (360TF), pseudo-3D layer in 360-Adapter (Pseudo-3D), and latitude-aware loss (Lat. Loss).}
    \label{fig:ablation}
    \vspace{-5pt}
\end{figure}

We primarily conducted ablation studies on the proposed 360 Text Fusion strategy, the pseudo-3D layer in the 360-Adapter, and the latitude-aware loss, as illustrated in Fig.~\ref{fig:ablation}. Given the prompt ``a car driving down a street next to a forest", the first row without 360TF can not generate the car because of low-quality captions in the training process. The second row without pseudo-3D layer can generate a car, but due to the lack of temporal modeling, the results exhibit flickering. The third row without latitude-aware loss can produce relatively good results, but it still falls slightly short in terms of clarity, field of view, and other aspects compared to the last row with the complete 360DVD.

\subsection{User Study}
31 participants were surveyed to evaluate the graphics quality, cross-frame consistency, left-right continuity, content distribution, and motion patterns of 8 sets of generated results. For each criterion, they selected the video they deemed most fitting for the theme of high-quality 360-degree panoramic videos. 
The data presented in Table~\ref{tab:web360} indicates that our model outperforms the other three methods significantly across all five dimensions. Simultaneously, our proposed 360ET can remarkably improve video quality, and left-right continuity, solely based on the native AnimateDiff and panoramic LoRA.

%% file: sec/5_conclu.tex
\section{Conclusion}
In this paper, we introduce 360DVD, a pipeline for controllable $360^{\circ}$ panorama video generation. 
Our framework leverages text prompts and motion guidance to animate personalized T2I models. 
Utilizing the proposed WEB360 dataset, 360-Adapter, and 360 Enhancement Techniques, our framework can generate videos that adhere to the content distribution and motion patterns in real captured panoramic videos. Extensive experiments demonstrate our effectiveness in creating high-quality panorama videos with various prompts and styles. 
We believe that our framework provides a simple but effective solution for panoramic video generation, and leads to inspiration for possible future works.